\documentclass{article}

\usepackage{arxiv}

\usepackage[utf8]{inputenc} % allow utf-8 input
\usepackage[T1]{fontenc}    % use 8-bit T1 fonts
\usepackage{hyperref}       % hyperlinks
\usepackage{url}            % simple URL typesetting
\usepackage{booktabs}       % professional-quality tables
\usepackage{amsfonts}       % blackboard math symbols
\usepackage{nicefrac}       % compact symbols for 1/2, etc.
\usepackage{microtype}      % microtypography
\usepackage{cleveref}       % smart cross-referencing
\usepackage{lipsum}         % Can be removed after putting your text content
\usepackage{graphicx}
\usepackage[numbers]{natbib}
\usepackage{doi}

\usepackage{adjustbox}
\usepackage{subcaption}

\title{Enabling and Understanding Personalization in AI-Generated Advertising Imagery}

\date{}

\newif\ifuniqueAffiliation
\ifuniqueAffiliation % Standard variant of author block

\else
\usepackage{authblk}

\author[1,2]{Victor Kolominsky-Rabas\thanks{\texttt{victor.kolominsky-rabas@fit.fraunhofer.de}}}
\author[1,2]{Leopold Müller}
\author[1]{Claudius Budcke}
\author[1]{\\Claas Christian Germelmann}
\author[1,2]{Niklas Kühl}

\affil[1]{University of Bayreuth, Universitätsstraße 30, 95447 Bayreuth, Germany}
\affil[2]{Fraunhofer FIT, Wittelsbacherring 10, 95444 Bayreuth, Germany}
\fi

\renewcommand{\headeright}{PREPRINT}
\renewcommand{\undertitle}{PREPRINT}
\renewcommand{\shorttitle}{Enabling and Understanding Personalization in AI-Generated Advertising Imagery}

\hypersetup{
pdftitle={Enabling and Understanding Personalization in AI-Generated Advertising Imagery},
pdfauthor={Victor Kolominsky-Rabas},
}

\begin{document}
\maketitle

\begin{abstract}
Personalized marketing traditionally matches static products to customers, while dynamic creative optimization focuses mainly on AI-driven text personalization or basic product image modifications. We address this gap by developing and implementing an AI-based framework that generates personalized advertising imagery directly from customer data. We evaluate this framework in a two-stage within-subject study with $N=100$ participants across four products and three levels of personalization, varied by the amount and specificity of customer data used. Participants rated each image on attitude toward the advertisement, attitude toward the product, and purchase intention. Results show that participants perceive differences across personalization levels and evaluate AI-generated advertising imagery most positively at a moderate level of personalization. High personalization increases perceived personalization, which is positively associated with all three outcome measures, but also increases perceived creepiness, which is negatively associated with the outcomes and dominates the total effect.
\end{abstract}

%\include{1_JBR/sections/0-abstract}

% keywords can be removed

\section{Introduction}
\label{sec:introduction}

Traditional digital personalized marketing uses customer data to optimize a single parameter: the match between a static product and a specific customer~\citep{chandra2022personalization,murray2009personalization,naumov2019deep}.
However, generative artificial intelligence (GenAI) introduces a second parameter, allowing the product's presentation to be dynamically generated for the viewer. If a customer who loves cycling but is not an environmentalist is in the market for a new electric vehicle, artificial intelligence (AI) can adapt the marketing material for the vehicle as a ``sports-utility tool'' rather than an ``eco-friendly car''. This shift in focus is enabled by the immense capabilities of modern GenAI systems. They have reopened the possibility of one-to-one communication---but at an unprecedented scale. For the first time, digital advertisements can be tailored not only to a segment but to the fine-grained details of a single individual.

Personalization is a powerful lever in digital marketing because it increases relevance, engagement, and profitability~\citep{chandra2022personalization}. Yet most personalization to date has focused on recommendation systems---that is, delivering an advertisement the customer is likely to be interested in---rather than tailoring the advertisement's content itself to the individual customer~\citep{smolinski2023towards}. The actual content of advertisements has typically been tailored only to broad market segments, while more granular tailoring has been constrained by the high costs of producing conventional marketing materials.

Literature shows that dynamic creative optimization (DCO) enables scalable personalization of advertising content by recombining predefined assets in real time~\citep{baardman2021dynamic,li2024two}, but it is limited to adjustable elements or small sets of preset images. Recent advances in message personalization demonstrate that large language models (LLMs) can generate persuasive, psychologically tailored text at scale~\citep{matz2024potential}. Related work shows that GenAI-based personalized video advertisements can increase engagement compared with personalized image ads and generic non-personalized video ads~\citep{kapoor2025frontiers}. However, this work focuses on personalized video messages based on purchase histories and click engagement, rather than on fully AI-generated advertising imagery, real customer profiles, and different depths of visual personalization. For advertising imagery, research remains narrow. A few studies generate advertising imagery, but most rely on inpainting that modifies backgrounds around a fixed product image~\citep{czapp2024dynamic,yang2024new} rather than creating new scenes in which the product is embedded and set in context. Research on synthetic advertising has highlighted that AI-based manipulation and generation technologies, such as deepfakes and generative adversarial networks, can create advertising content automatically and may enable new forms of hyperpersonalization~\citep{campbell2022preparing}. Moreover, emerging empirical work has begun to examine how customers respond to AI-generated advertisements, for example in the context of disclosure and luxury-brand authenticity~\citep{to2025ai}, while broader marketplace-level research shows that GenAI affects the production, consumption, and substitution of creative goods such as stock imagery~\citep{goldberg2025generative}. However, these studies do not investigate how GenAI can be used to personalize advertising imagery from real individual customer data, nor how different levels of such personalization affect marketing outcomes. To date, no study has systematically examined the use of GenAI for creating personalized advertising imagery, based on real customer data, or varied the level of personalization. It is unclear a) how such personalization would be implemented on a technical level and b) whether such personalization is beneficial for marketing purposes. To address this, we derive two research questions:

\begin{itemize}
    \item[RQ1] \textit{How can we design a technical framework providing AI-generated personalized advertising imagery?}
    
    \item[RQ2] \textit{How does the level of personalization in AI-generated advertising imagery influence attitude toward the advertisement, attitude toward the product and purchase intention?}
\end{itemize}

\begin{figure}[ht]
  \begin{center}
  \includegraphics[width=\linewidth]{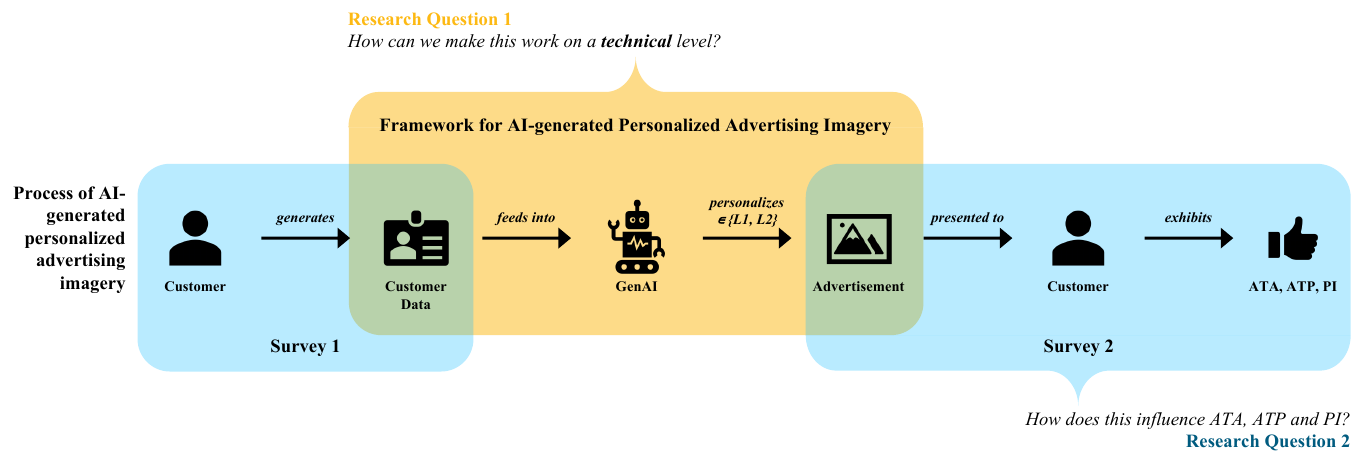}
  \end{center}
  \caption{Overview of the research design along the process of AI-generated personalized advertising imagery. To answer RQ1, we develop and implement a technical framework that transforms customer data into personalized advertising imagery at different levels of personalization (L0--L2). This framework is a prerequisite for answering RQ2, as it produces the advertisements evaluated in our two-stage study: Survey~1 captures participant data as framework input, and survey~2 presents the resulting images to the same participants and measures ATA, ATP and PI.}
  \label{fig:overview}
\end{figure}

\Cref{fig:overview} provides an overview of our research design across the process of AI-generated personalized advertising imagery, linking the technical framework (RQ1) to the empirical evaluation of attitude toward the advertisement (ATA), attitude toward the product (ATP) and purchase intention (PI) at different levels of personalization (RQ2).
To answer RQ1, we examine current manual workflows for creating traditional advertising imagery.
Inspired by this, we develop a technical framework for AI-generated personalized advertising imagery and implement it combining a widely accessible, state-of-the-art LLM with a text-to-image (T2I) generator.
Personalization is implemented across three levels (L0--L2) that range from a non-personalized baseline to hyperpersonalized outputs---realized by varying the amount and specificity of customer data fed into the framework.
To address RQ2, we develop three hypotheses: H1 tests whether personalization affects ATA, ATP and PI across L0--L2, H2 tests improvements relative to the non-personalized baseline, and H3 tests for an inverted-U relationship in which moderate personalization is most effective while excessive tailoring reduces effectiveness.
We conduct a two-stage study (consisting of survey~1 and survey~2) with a within-subjects design of $N=100$ participants using four exemplary products.
Survey~1 collects real participant data used as input for the implemented framework for AI-generated personalized advertising imagery.
The generated personalized advertising imagery is evaluated in survey~2 using established items to measure ATA, ATP and PI.

Results show that participants perceive differences in the levels of personalization, which confirms our assumption that GenAI is able to incorporate customer data into generated advertising imagery and vary its level of personalization. Across ATA, ATP, and PI, light personalization (L1) produces the most favorable responses. The stronger level of personalization (L2) does not add additional benefit and can reduce ATA, ATP and PI. Confirmatory analyses support H1 for all three main constructs. H2 is only partially confirmed for ATA and ATP (through positive impact at L1 only) and tests show significant effects for ATA and ATP for H3 as well. An exploratory analysis reveals an underlying mechanism in which perceived personalization and perceived creepiness strongly influence ATA, ATP and PI. In summary, this work contributes to the state-of-the-art in four ways.

\begin{enumerate}
    \item We introduce a technical framework, along with its implementation, that couples an LLM with a T2I generator to produce advertising imagery from real customer data and demonstrate its feasibility.
    \item Our work provides a controlled within-subjects evaluation based on real participant profiles at the image level and tests effects on ATA, ATP and PI complemented by manipulation checks of perceived personalization and perceived creepiness.
    \item We offer evidence that light personalization outperforms stronger personalization, which aligns with the personalization-paradox~\citep{aguirre2015unraveling} and yields practical guidance for the low-cost, scalable deployment of personalized image-based advertising with GenAI.
    \item We propose an explanation for the diminishing returns of personalization in AI-generated personalized advertising imagery by connecting perceived creepiness with limitations of current GenAI model capabilities.
\end{enumerate}

The remainder of this paper is organized as follows. \Cref{sec:related_work} reviews the relevant literature on personalized marketing and GenAI, establishing the theoretical foundation. \Cref{sec:research_dev} describes the research design and implementation of the technical framework, as well as the development of the hypotheses. \Cref{sec:study_design} presents the study design for survey 1 and survey 2, with the results reported in \Cref{sec:results}. Finally, \Cref{sec:discussion} discusses and interprets the results, addresses theoretical and practical implications, and suggests directions for future research. \Cref{sec:conclusion} concludes the work.
\section{Background \& related work}
\label{sec:related_work}

Personalization in marketing aims to enhance the customer experience and increase revenue by making advertisements more relevant to individual customers~\citep{chandra2022personalization,kim2022getting}. Higher relevance can increase attention, clicks, and purchases~\citep{matz2017psychological,dekeyzer2022and,bang2019level}. At the same time, personalization can become counterproductive when it is perceived as obtrusive or privacy-invasive~\citep{goldfarb2011online,malheiros2012too}. This tension is known as the personalization-paradox: using more customer data can increase relevance, but it can also trigger privacy concerns and reactance~\citep{aguirre2016personalization,brinson2016juxtaposing}. Understanding how different forms and degrees of personalization affect customer responses is therefore central to personalized marketing.

\paragraph{From recommendation to creative personalization.}
A large body of work optimizes allocation by matching existing advertisements to users through recommendation systems \citep{chandra2022personalization,murray2009personalization,naumov2019deep}. This improves who sees which advertisements, while the content itself often remains template-based.
Modern GenAI provides two pillars for the shift to personalized content. Transformer LLMs enable controllable text generation and prompt construction~\citep{brown2020language}. Diffusion models enable high-fidelity image synthesis from text prompts~\citep{rombach2022high, betker2023improving}. Together, they support scalable, automated pipelines that shift from showing existing advertisements to the right person to generating a personalized advertisement for any given person.

\paragraph{Dynamic creative optimization.}
DCO automates the assembly of creatives from modular assets, which enables large-scale adaptation while keeping production costs manageable~\citep{li2024two}. In practice, services such as Facebook's dynamic ads~\citep{meta_dynamic_ads} operationalize this approach by linking product catalogs to user intent signals and composing variants at impression time to improve relevance and click-through-rate (CTR). Academic work frames the selection of products and creatives as a joint optimization problem and shows that algorithmic policies can outperform heuristic, rule-based settings that are still common~\citep{baardman2021dynamic}. Early generative systems illustrate a transition point. Instead of only recombining assets, like the previous approaches, they start to produce tailored visuals from predicted user attributes, for example, personality-driven imagery generated with StyleGANs~\citep{farseev2021somin}. The key limitation remains that most DCO pipelines select or lightly adapt predefined elements, rather than create completely new scenes that reflect highly individual user details.

\paragraph{GenAI-driven message personalization.}
Conceptual work positions GenAI as a driver of hyper-personalization across text, image, video, and audio~\citep{patil2024generative}. Empirical studies show that LLMs can generate persuasive messages that align with user characteristics~\citep{lee2024developing}. Large-scale experiments demonstrate that psychologically tailored messages outperform generic or mismatched messages, with effects that persist even when the AI origin is known~\citep{matz2024potential}. These findings establish that GenAI can craft effective, personalized text at scale, motivating testing of whether analogous benefits carry over when personalization is enacted visually.

\paragraph{Customer responses to AI-generated advertising.}
A related stream examines synthetic and AI-generated advertising more broadly. \citet{campbell2022preparing} provide a framework for understanding customer responses to manipulated and synthetic advertising, including deepfakes and AI-generated ads. Recent empirical work shows that AI-generated advertisements can shape customer responses, for example through disclosure effects and perceived authenticity in luxury advertising~\citep{to2025ai}. Beyond advertising, marketplace-level research shows that GenAI affects creative goods markets such as stock imagery by changing production, variety, sales, and substitution between GenAI and non-GenAI content~\citep{goldberg2025generative}. These studies establish the relevance of AI-generated advertising and imagery, but they do not examine how advertising images can be personalized from real individual customer data or how different levels of such personalization affect marketing outcomes.

\paragraph{GenAI-driven image personalization.}
Research on GenAI-driven image personalization in marketing remains nascent and heterogeneous. Several systems chain an LLM with a T2I model to automate prompt design for banners and product visuals, although personalization often refers to better product matching rather than user-specific scenes~\citep{vashishtha2024chaining}. Other pipelines keep the product pixels fixed and personalize the surrounding context through inpainting or controlled background changes, which can increase online CTR in field tests~\citep{czapp2024dynamic,yang2024new}. Brand-alignment approaches fine-tune diffusion models to match brand personality and funnel objectives, but they operate primarily at the brand level rather than the individual level~\citep{jansen2023automated}. Comparative studies show that GenAI-generated images can rival or surpass human-created images in quality and performance, including real-world CTR, but the generated content is typically generic rather than tailored to individual user profiles~\citep{hartmann2025power}. Where personalization appears, it often relies on coarse segments or narrow psychographic splits, such as introversion versus extraversion, and evaluations sometimes use automated semantic similarity measures such as CLIP instead of human outcome measures~\citep{smolinski2023towards,shilova2023adbooster,radford2021learning}.

Across these strands of research, prior work provides important building blocks but leaves three limitations. First, recommendation-based personalization and DCO mostly optimize allocation or recombine predefined assets rather than generate fully novel advertising scenes grounded in individual profiles~\citep{murray2009personalization,naumov2019deep,chandra2022personalization,li2024two,baardman2021dynamic}. Second, GenAI-based personalization has been studied mainly for text, peripheral image edits, brand-level generation, or coarse segments rather than individualized visual narratives~\citep{matz2024potential,czapp2024dynamic,yang2024new,jansen2023automated,vashishtha2024chaining}. Third, evaluations often prioritize CTR, marketplace outcomes, or automated similarity measures instead of user-centered advertising effectiveness constructs such as ATA, ATP, and PI~\citep{hartmann2025power,smolinski2023towards,radford2021learning}. Thus, to the best of our knowledge, no study has systematically investigated how GenAI-based personalization of advertising imagery affects customer responses. In particular, prior work has not tested fully AI-generated personalized advertising imagery that embeds products into customer-tailored visual narratives, uses real individual customer data, explicitly varies the depth of personalization, and evaluates its effects on established marketing outcomes.
\section{Research development}
\label{sec:research_dev}

In this section, we introduce the technical framework for AI-based personalized advertising imagery and its implementation to answer RQ1. We then develop the hypotheses to address RQ2, where the framework is applied.

\subsection{AI-based personalized advertising imagery: Framework and implementation}
\label{sec:image_generation}

First, we conceptualize the framework, by defining relevant steps along the creative workflow in advertising~\citep{butterfield2009excellence} and then link the conceptual framework to its technical implementation, aiming for full automation. We then discuss how a controlled variation in personalization is achieved.

\subsubsection{Conceptual framework for AI-generated personalized advertising imagery}
\label{sec:framework}

The framework aims to generate personalized advertising imagery by embedding products within customer-tailored visual narratives, i.e. images that combine visual cues relatable to the target customer with the product in a familiar environment. The framework is designed to address RQ1. To achieve this objective in a realistic scenario, where each individual customer can be addressed, the framework must fully automate all steps that vary between customers so that the system can scale.

\begin{figure}[tbh]
  \begin{center}
  \includegraphics[width=0.65\linewidth]{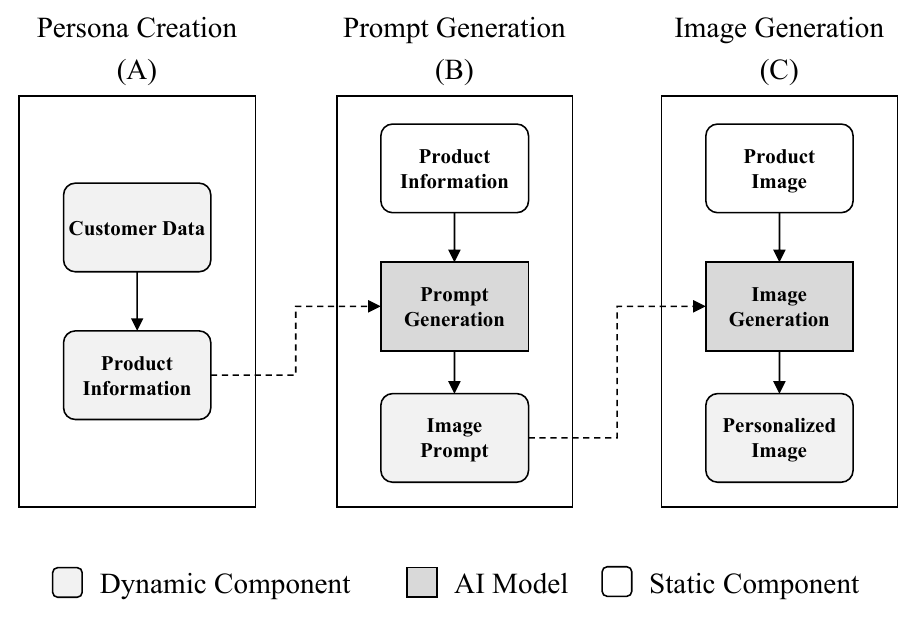}
  \end{center}
  \caption{The framework for personalized image generation consists of three central components: the persona creation (A), the image prompt generation (B), and the generation of the personalized image (C).}
  \label{fig:framework}
\end{figure}

The framework consists of three sequential steps (A, B, C), illustrated in \Cref{fig:framework}. The three steps are derived from the conventional creative workflow, which consists of strategy development, creative briefing, creative work, and campaign roll-out~\citep {butterfield2009excellence,kotlerMarketingManagement2016}. Creative workflow typically begins with strategy development, e.g., identifying the target audience. The creative workflow then proceeds to creative briefing, which defines the conceptual properties the final advertisement should convey. This is followed by creative work, the actual production of marketing content, and finally campaign roll-out \citep{butterfield2009excellence,kotlerMarketingManagement2016}. Within the framework, key steps of the creative workflow are automated. Strategy development commonly includes the creation of personas: detailed fictitious customer profiles that portray a typical customer derived from preceding market research. The framework adapts this step, but instead of generalized personas, the framework creates an individual persona description derived directly from the customer data of a single individual. The framework also replaces the conventional processes of preparing a creative brief and executing creative work with two distinct AI models. The created persona description is combined with fixed product information and processed by the first AI model, an LLM, to generate a personalized image prompt that specifies in detail how the final image should look and feel. We see this image prompt in step prompt generation as analogous to the creative brief produced by a marketing team, which would typically be forwarded to a design agency. In the final step, image generation, we automate the creative work and the image prompt, together with a reference image of the product, is passed to the T2I model, which uses it to generate the final hyperpersonalized advertising imagery.
With our framework, all dynamic components that vary across customers are fully automated, enabling the generation of personalized advertising images for any number of customers and products.

\subsubsection{Technical implementation of the framework for AI-generated personalized advertising imagery}
\label{sec:personalization_levels}

\begin{figure}[tbh]
  \begin{center}
  \includegraphics[width=0.8\linewidth]{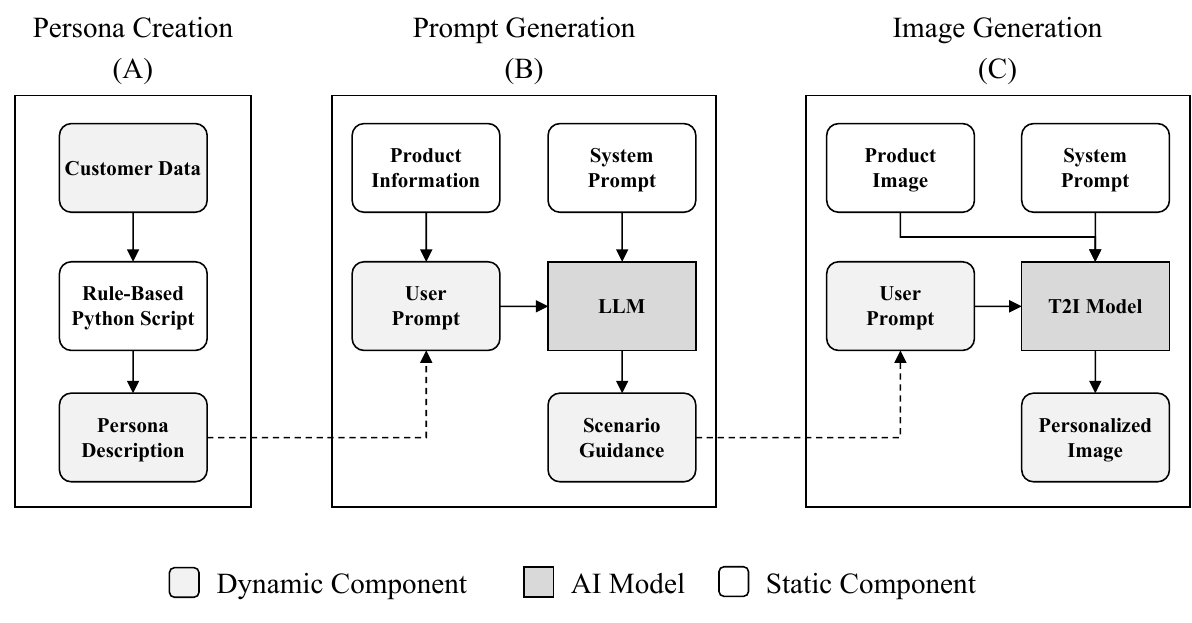}
  \end{center}
  \caption{The technical implementation of the framework for AI-generated personalized advertising imagery operationalizes the three steps of the framework with a rule-based Python script for persona creation (A), an LLM for scenario guidance generation (B), and a T2I model for generating the personalized image (C).}
  \label{fig:pipeline}
\end{figure}

The conceptual framework is implemented in Python, depicted in \Cref{fig:pipeline}.%, and is available at \url{https://anonymous.4open.science/r/GenAI_Personalized_Advertisements_Framework}. 
It executes the three steps introduced above. In step persona creation (A), the structured customer data is processed by a rule-based Python script that embeds the values into predefined natural-language text snippets. An example of such a generated persona description is: \textit{``The persona is a female individual, 35 years old, born in the United States, currently living in Bakersfield, United States.''}
In step prompt generation (B), the previously created persona description is inserted into a predefined user prompt template, which serves as input to the LLM. This template also includes placeholders for product information, which are dynamically filled at runtime, such as the product name and product type, e.g. ``Polestar 4'' and ``electric vehicle''. The resulting user prompt is then combined with a system prompt and passed to an LLM, operationalized through OpenAI's model gpt-5. From there, the LLM generates a natural-language description of the intended personalized visual scenario, following the instructions in the system and user prompts, which include the persona description and product information. This image description provides scenario guidance for the final advertising image.
In step image generation (C), the previously generated scenario guidance is inserted into a predefined user prompt template tailored for the T2I model. This step is implemented using OpenAI's gpt-image-1 model, which additionally accepts a reference image of the advertised product to ensure accurate visual depiction. Alongside the model, prompt, and reference image, additional parameters such as image quality (high) and resolution (1024x1536) are specified. The T2I model then transforms these inputs into the final personalized advertising image through a one-shot generation process.

\paragraph{Prompt template design.}

The prompt templates serve as the central control mechanism of the implemented framework. They keep creative instructions constant across products and personalization levels while allowing product- and customer-specific variation through dynamically filled placeholders. The implementation uses separate prompt templates for the LLM and the T2I generator, as shown in \Cref{fig:pipeline}, with placeholders for product name, product type, and persona description. To ensure consistent outputs, the templates use techniques such as role prompting, Markdown formatting, XML tags, and negative prompts to prevent unwanted textual elements such as slogans or graphic overlays~\citep{phoenix2024prompt,muller2025data}.

\paragraph{Personalization.}

To isolate the central manipulation, the prompt templates remain identical across all levels of personalization. Variation is introduced only through the persona description created in the persona creation step. For L0, no customer data are included, making it the non-personalized baseline. For L1 and L2, the rule-based Python script creates persona descriptions that differ in the amount and specificity of customer data. The LLM then translates these persona descriptions into scenario guidances, which are used by the T2I model to generate images with increasing levels of personalization.
The rationale is that customer data can be ordered along a continuum from general to highly individual information and grouped into increasingly specific market segments, up to the market of one. Since templates remain constant across levels, systematic differences in the resulting images stem from the customer data used as input, apart from inherent variation in GenAI outputs. Personalization is therefore not imposed directly at the image level, but emerges from the amount and specificity of the customer data. This setup isolates the intended manipulation and enables a controlled test of how different degrees of personalization influence ATA, ATP, and PI.

\subsection{Hypothesis development}
\label{sec:hypothesis_dev}

In this section, we outline the conceptual foundation of the empirical study, specifying what is being investigated and why. It defines the study's independent and dependent variables, introduces exploratory constructs, and explains how marketing communication success is operationalized.

\subsubsection{Study variables}
\label{sec:study_variables}

To answer RQ2, the independent variable is the level of personalization, operationalized through differently personalized images. To measure the effect of the levels of personalization, we use three different constructs. In addition, exploratory variables enable further investigation into factors that may shape or moderate responses to AI-generated personalized content.

\paragraph{Independent variable: Level of personalization.} 
The independent variable is the level of personalization, operationalized through three advertising images per product generated by the framework described above. The levels range from L0 to L2: L0 is a generic, non-personalized baseline without customer data; L1 uses general demographic and geographic information; and L2 adds more specific sociodemographic, lifestyle, household, pet ownership, and appearance-related attributes. Thus, the conditions differ only in the amount and specificity of customer data used as input. The levels are conceptually defined rather than equidistant and follow a logic of increasing individualization from general to highly specific customer information~\citep{wilkieConsumerBehavior1994}. All included customer information is relevant for conventional market segmentation and is either currently collected or technically feasible to collect in digital marketing contexts~\citep{kotlerMarketingManagement2016,googlePrivacyPolicyPrivacy2025,baardman2021dynamic}.

\paragraph{Dependent variables: Attitude toward the ad, attitude toward the product and purchase intention.}
The dependent variables in this study reflect ATA, ATP and PI, key marketing outcome constructs commonly used in advertising research~\citep{morrison1979purchase,bruner2005marketing,lutz1985affective}. We select three constructs for confirmatory hypothesis testing due to their widespread use and empirical validation as indicators of advertising effectiveness. Each offers a unique perspective on customer response:

\begin{itemize}
\item \textbf{Attitude toward the advertisement} captures participants' affective evaluation of the advertisement as a stimulus, reflecting a predisposition to respond positively or negatively to the advertisement during a specific exposure occasion~\citep{lutz1985affective}.
\item \textbf{Attitude toward the product} captures both affective and cognitive evaluations of the advertised offering, representing the customer's overall evaluative judgment of the product~\citep{bruner2005marketing}.
\item \textbf{Purchase intention} captures the customer's stated likelihood of buying the advertised product, representing a forward-looking indicator of behavioral intent~\citep{morrison1979purchase}.
\end{itemize}

The order of these measures reflects a hierarchy rooted in the widely acknowledged hierarchy-of-effects model proposed by \citet{lavidge1961model}.

\paragraph{Exploratory variables.}
\label{sec:exploratory_variables}

In addition to the dependent variables used for hypothesis testing, this study includes exploratory measures to enrich the interpretation of participants' perceptions and responses to AI-generated personalized images.

\begin{itemize}
    \item \textbf{Perceived personalization}: Assesses whether participants recognize the intended level of personalization (L0–L2).

    \item \textbf{Perceived creepiness}: Excessive personalization may evoke feelings of creepiness, which is associated with privacy concerns and a sense of being overly monitored~\citep{moore2015creepy,petrova2025phenomenon,krause2026technically,goldfarb2011online}. In AI-generated images, perceptual dissonance can arise when stimuli appear human-like but not fully convincing~\citep{brink2019creepiness}. This reflects the uncanny valley effect, where images that are almost but not fully human elicit discomfort and aversion~\citep{draude2011intermediaries}.
\end{itemize}

\subsubsection{Confirmatory hypotheses}
\label{sec:hypotheses}

We derive three confirmatory hypotheses to answer RQ2. They investigate the effect of image personalization on ATA, ATP and PI, and rest on theoretical and empirical foundations.

Advertising images generated by GenAI can perform on par with or outperform human-designed imagery \citep{hartmann2025power,jansen2023automated}. Prior work shows positive effects of personalization from recommendation~\citep{pathak2010empirical}, through DCO~\citep{baardman2021dynamic}, to language-model personalization and GenAI-based image personalization~\citep{czapp2024dynamic,vashishtha2024chaining}. Given the novelty of GenAI-driven image personalization, hypothesis 1 (H1) tests for any effect, and hypothesis 2 (H2) examines the expectation of an overall positive effect.

\begin{itemize}
\item [\textbf{H1}] There is an effect of image personalization on marketing communication success. \textit{Non-directional}. At least one mean outcome across the three levels of personalization (L0--L2) differs.
\item [\textbf{H2}] There is a positive effect of image personalization on marketing communication success. \textit{Directional}. At least one mean outcome across the three levels of personalization (L0--L2) differs in a positive direction.
\end{itemize}

Evidence for a positive effect includes field experiments, where matching appeals to personality traits increased clicks by up to 40\% and purchases by up to 50\%~\citep{matz2017psychological}. GenAI-enabled message personalization improves engagement across domains~\citep{matz2024potential}. In display advertising, higher content personalization raises CTR early in the decision process~\citep{bleier2015personalized}. Tailored offerings elicit favorable attitudes~\citep{goldsmith2004have}, and micro-level cues such as a customer's name strengthen attention and memorability~\citep{abdel2021effectiveness}. For visual content, AI-generated creatives can align with customer preferences and facilitate conversions, though effects on perceived authenticity can be mixed~\citep{sharma2024user, nazrin2025ai}.

Personalization may also reach a tipping point. Reactance arises when advertisements expose uniquely identifiable details~\citep{malheiros2012too}, and greater data usage can heighten privacy concerns~\citep{aguirre2016personalization, brinson2016juxtaposing}. Perceived creepiness rises from low to moderate personalization and remains elevated at high levels~\citep{de2022going}. AI-generated visuals can face tensions between appeal and authenticity~\citep{sharma2024user, nazrin2025ai}. Hyper-realistic faces are vulnerable to uncanny valley effects when subtle imperfections occur~\citep{draude2011intermediaries}, which is relevant for higher levels of personalization where human figures are generated from customer-specific traits. Stylized or cartoon-like visuals can avoid this by lowering expectations for realism~\citep{draude2011intermediaries}. Together, these findings suggest a ``sweet spot'' of personalization. Therefore, we define hypothesis 3 (H3):

\begin{itemize}
\item [\textbf{H3}] The relationship between image personalization and marketing communication success follows an inverted-U pattern. \textit{Directional, non-linear}. The moderate level of personalization (L1) enhances marketing outcomes most effectively, while excessive personalization (L2) may reduce effectiveness due to customer resistance or perceived intrusiveness.
\end{itemize}

\subsubsection{Pre-registration and ethical approval}
\label{sec:pre-registration_irb}

To ensure transparency and methodological rigor, the study was preregistered on the \textit{Open Science Framework} (OSF) prior to data collection. The preregistration was created and registered on August 20, 2025. The full preregistration can be accessed here: \url{https://osf.io/uz5je/overview?view_only=2907997569564d73b62de212cd048d47}.
Additionally, the study received ethical approval from the Ethics Committee of the University of Bayreuth on May 28, 2025, prior to data collection. All participants provided informed consent and were compensated in accordance with Prolific's~\citep{prolific} ethical guidelines.
\section{Study design}
\label{sec:study_design}

We now turn to the study's implementation: how we select products, measure our dependent and exploratory variables, design the two surveys, and recruit participants.

\subsection{Product selection}
\label{sec:product_selection}

For the study, we need to select products to include in the advertising imagery. The selection follows three criteria to ensure a diverse yet balanced set that allows for generalizable insights into AI-generated personalized advertising. First, we apply the FCB Grid~\citep{vaughn1980advertising} as a theoretically grounded framework. The FCB Grid distinguishes products along the degree of involvement in the purchase decision (low vs.\ high) and the dominant decision driver (thinking vs.\ feeling)~\citep{vaughn1980advertising}. We select one product from each quadrant to cover different types of purchase decisions. Second, the selection seeks to balance potential demographic interests by combining products that stereotypically attract male, female, and gender-neutral interests. Third, expected visual portrayals generated by the framework are considered, aiming to balance likely indoor and outdoor depictions across products.

The reference images used in the image generation step are obtained from publicly available product images and are cited in \Cref{tab:reference_images}. We select the Polestar 4 as an informative high-involvement product, the RIMOWA Original Cabin as an affective luxury product, Tide Ultra Hygienic Clean as a habitual everyday product, and Lindt Lindor Truffles as a self-satisfaction product~\citep{cheong2017revisiting,teng2010use,glowa2002white}.

\begin{table}[htbp]
\centering
%\resizebox{\linewidth}{!}{
\begin{tabular}{lll}
\toprule
FCB quadrant & Product & Product category \\
\midrule
Informative & Polestar 4~\citep{polestar2024} & Electric vehicle \\
Affective & RIMOWA Original Cabin~\citep{rimowa2026} & Luxury suitcase \\
Habitual & Tide Ultra Hygienic Clean~\citep{tide2020} & Laundry detergent \\
Self-satisfaction & Lindt Lindor Truffles~\citep{lindt2013} & Chocolates \\
\bottomrule
\end{tabular}
%}
\caption{Reference products used in the image generation step. For copyright reasons, the original product images are not reproduced in the manuscript.}
\label{tab:reference_images}
\end{table}

\subsection{Measurement of dependent variables}
\label{sec:dv_measures}

All three constructs (ATA, ATP and PI) are measured on a five-point Likert scale with single-item adaptations from validated multi-item scales to reduce participant burden~\citep{bergkvist2007predictive,fuchs2009using}. The order of measurement follows the hierarchy-of-effects model~\citep{lavidge1961model}, progressing from conative to affective to stimulus-focused evaluation. ATA is based on the five-item scale by \citet{lee1999responses} as reported by~\citet{bruner2005marketing}. ATP draws on the seven-item scale originating from \citet{shamdasani2001location} as cited in~\citet{bruner2005marketing}. PI uses items summarized by~\citet{bruner2005marketing} that build on work by \citet{bower2001highly} and \citet{bower2001beauty}. The exact single-item wordings are listed below. The product description for PI is adjusted, depending on the specific product.

\begin{itemize}
    \item \textbf{PI}: ``If I were looking for an electric vehicle, based on this image, I would consider purchasing this product.''
    \item \textbf{ATP}: ``This image makes me feel positive toward the product.''
    \item \textbf{ATA}: ``The advertisement image is appealing to me.''
\end{itemize}

\subsection{Measurement of exploratory variables}
\label{sec:ev_measures}

All exploratory constructs are measured with concise single-item formulations on a five-point Likert scale, adapted to the context of AI-generated personalized images where applicable. Perceived personalization is adapted from \citet{kalyanaraman2006psychological}. Perceived creepiness condenses the multi-item semantic differential by \citet{de2022going} and reflects uncanny responses in near-human visuals~\citep{brink2019creepiness, draude2011intermediaries}.
The exact wordings are listed below.

\begin{itemize}
    \item \textbf{Perceived personalization}: ``The scene featured in the image targets me as a unique individual.''
    \item \textbf{Perceived creepiness}: ``I find this advertisement image creepy.''
\end{itemize}

\subsection{Survey implementation}
\label{sec:survey_implementation}

The study is conducted using two online surveys. In survey~1, participants provide personal information that serves as input for the generation of personalized advertisement images. In survey~2, these images are presented back to the identical participants, who then evaluate them across the dependent and exploratory variables described in the preceding sections. Both surveys are created and administered with LimeSurvey hosted on a department-managed server to ensure data security and compliance with institutional requirements. 
%The experiment follows a within-subjects design.

\subsubsection{Survey 1: Data collection}
\label{sec:SOUP-1}

Survey~1 is designed to collect participant details that are subsequently used for generating personalized advertising imagery.

\paragraph{Data collection criteria.}
To decide which details are collected from the participants in survey~1, three criteria are applied to justify their inclusion. (1) The data point must be relevant for image generation. (2) The data point must be relevant for market segmentation. (3) The details collected must be, at least in theory, collectible in real-world practice.

\paragraph{Structure of survey 1.} 
Survey~1 collects participant information used as input for personalized image generation and is divided into five question groups: \emph{basic information}, \emph{life stage}, \emph{lifestyle}, \emph{environment}, and \emph{style}. Basic information covers \emph{gender}, \emph{age}, \emph{country of residency}, \emph{city of residency}, \emph{country of birth}, \emph{ethnicity}, and \emph{family origin/background}. Life stage captures \emph{occupation type}, \emph{occupation title}, \emph{income class}, \emph{relationship status}, \emph{number of children}, and \emph{age groups of children}. Lifestyle collects \emph{regular activities and hobbies} as free-text input to support realistic scenario generation. Environment covers \emph{housing type}, \emph{residence area}, \emph{climate}, \emph{attention check}, and \emph{pet ownership} with \emph{pet type} if applicable. Finally, style captures \emph{hair style}, \emph{hair color}, \emph{beard style} where applicable, \emph{body type}, \emph{glasses wearer}, \emph{color preference}, and \emph{style preference}. These variables are included because they are relevant for image generation, correspond to established segmentation dimensions, and are in principle collectible or inferable in digital marketing contexts~\citep{kotlerMarketingManagement2016,googlePrivacyPolicyPrivacy2025,wilkieConsumerBehavior1994,czinkota2021market,li2016human,baardman2021dynamic,adilova2024personalized}.

\subsubsection{Survey 2: Image evaluation}
\label{sec:SOUP-2}

\begin{figure}[htbp]
    \centering
    \begin{minipage}[t]{0.24\textwidth}
        \centering
        \includegraphics[width=\textwidth]{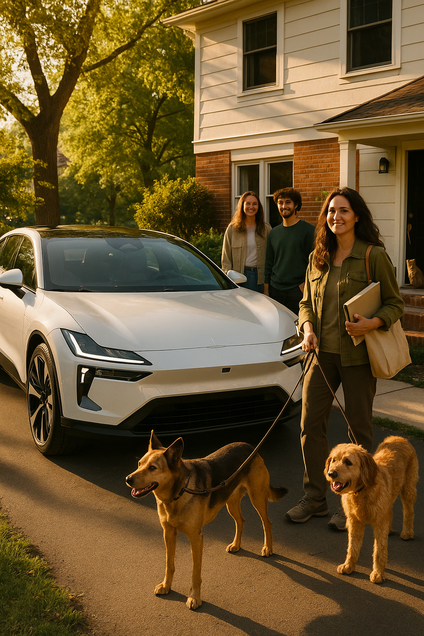}
    \end{minipage}
    \hfill
    \begin{minipage}[t]{0.24\textwidth}
        \centering
        \includegraphics[width=\textwidth]{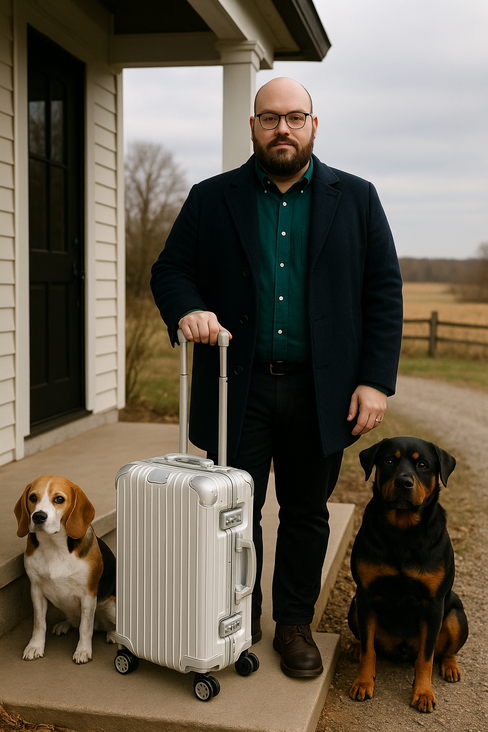}
    \end{minipage}
    \hfill
    \begin{minipage}[t]{0.24\textwidth}
        \centering
        \includegraphics[width=\textwidth]{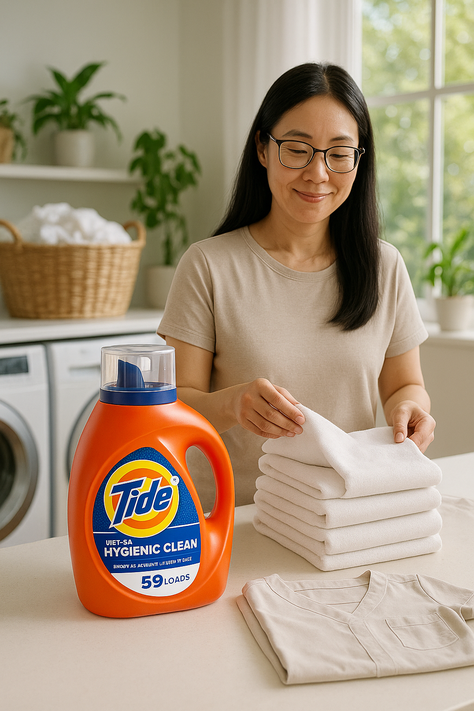}
    \end{minipage}
    \hfill
    \begin{minipage}[t]{0.24\textwidth}
        \centering
        \includegraphics[width=\textwidth]{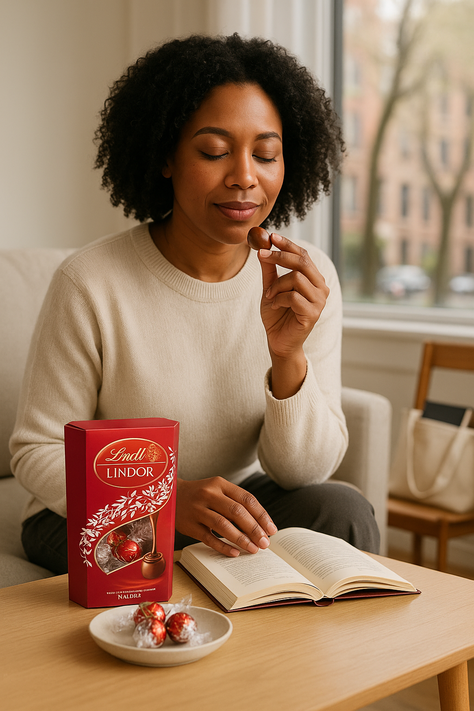}
    \end{minipage}
    
    \medskip
    
    \begin{minipage}[t]{0.24\textwidth}
        \centering
        Electric vehicle
    \end{minipage}
    \hfill
    \begin{minipage}[t]{0.24\textwidth}
        \centering
        Luxury suitcase
    \end{minipage}
    \hfill
    \begin{minipage}[t]{0.24\textwidth}
        \centering
        Laundry detergent
    \end{minipage}
    \hfill
    \begin{minipage}[t]{0.24\textwidth}
        \centering
        Chocolates
    \end{minipage}
    
    \caption{Fully AI-generated advertising imagery with the highest level of personalization (L2) for all products over four different participant personas.}
    \label{fig:example_images_good}
\end{figure}

In survey~2, participants evaluate the AI-generated advertisement images. A selection of four AI-generated advertisements from the survey are shown in \Cref{fig:example_images_good}. No generated images are manually filtered, replaced, or excluded before survey~2. Thus, participants evaluate the complete output of the automated generation pipeline, including occasional visual artifacts or low-quality generations.

\paragraph{Structure of survey 2.}
First, for each product, participants rate all three levels of personalization (L0--L2) on a fixed set of five items. These include the three dependent variables used for hypothesis testing (PI, ATP, and ATA) and two exploratory items: a manipulation check of perceived personalization and a measure of perceived creepiness. 
Therefore, each product evaluation section contains $5$ items $\times$ $3$ images = $15$ Likert ratings. After completing a product evaluation section, participants proceed to the next until all four products are evaluated. The order of the product evaluation pages is randomized to avoid biased responses from product order effects. Items within each product evaluation section are deliberately not randomized, as they are commonly presented in the sequence suggested by the hierarchy-of-effects model~\citep{lavidge1961model}. Participants were not explicitly informed before or during the image evaluation tasks that the presented advertisement images were generated using AI. This is important because prior work shows that perceived or communicated AI involvement can itself influence evaluations of generated content~\citep{knight2023generative,to2025ai}.

\paragraph{Evaluation of images.}
The evaluation setup is identical across all product evaluation sections, with only the product name and corresponding images varying.
Each evaluation is presented in a tabular layout: the leftmost column displays one image per row, followed by the five-point Likert response options (strongly disagree to strongly agree). Images are always displayed at full size, so each image is always fully visible. The question text remains fixed directly beneath the navigation bar, and the Likert labels stay anchored at the top of the table. This guarantees that evaluation criteria are always visible, reducing cognitive load and supporting accurate responses. The layout is optimized with the feedback from several pre-studies.

\subsection{Recruitment strategy}
\label{sec:recruitment}

Participants are recruited via Prolific~\citep{prolific}.
Participation is restricted to desktop devices to safeguard data quality~\citep{schlosser2018mobile,andreadis2015web}. The introduction page of each survey states the terms and conditions, the use of attention checks, and the consequences of failing them. Screening in survey~1 requires participants to be between 18 and 99 years old, native English speakers, and located in the United States. For survey~2, no additional screening is used, since access is limited to the saved group from survey~1. The sample distribution mode is set to ``standard''.
\section{Results}
\label{sec:results}

The following section presents the sample characteristics and the study's empirical findings.

%%% SUBSECTION %%%
\subsection{Sample characteristics}
\label{sec:sample_description}

The final sample consists of $N = 100$ participants. In total, 122 individuals initially submit responses for survey 1, with 3 excluded due to failed attention checks. From the remaining 119 individuals, 107 return to survey 2. After 7 further exclusions due to failed attention checks, 100 participants are included in the final analyses. Participants rated images across three levels of personalization over four different products resulting in a total of $1{,}200$ ratings. Participants have a mean age of $44.5$ years ($SD = 13.1$), with ages ranging from 22 to 84 years. The gender distribution is balanced, with 53 male (53\%) and 47 female (47\%) participants. Zero participants selected the ``diverse'' option. The large majority are U.S. citizens (92 participants, 92\%), in line with the screening criteria. Ethnic background is reported as 79 White (79\%), 8 Mixed (8\%), 6 Asian (6\%), 5 Black (5\%), and 2 Other (2\%). The final sample consists exclusively of native English speakers. 75 participants (75\%) indicate working either full-time, part-time, or self-employed, 13 participants (13\%) report being unemployed, 9 participants (9\%) report being retired, and 3 participants (3\%) report being university students.
The median completion time for survey~1 is $4.6$ minutes ($M = 5.2$, $SD = 2.6$, range $= 1.8$--$18.9$). 
The median completion time for survey~2 is $9.8$ minutes ($M = 10.4$, $SD = 4.1$, range $= 3.8$--$23.8$).
Compensation is in line with Prolific's guidelines for fair hourly payment at \pounds$8.57$/hr for survey~1 and \pounds$8.47$/hr for survey~2. No major technical issues were observed during the study. %The data and code for analysis are available at \url{https://anonymous.4open.science/r/GenAI_Personalized_Advertisements_Analysis}.

\subsection{Analysis}
\label{sec:descriptive_stats}
\begin{figure}[tbh]
  \begin{center}
  \includegraphics[width=0.7\linewidth]{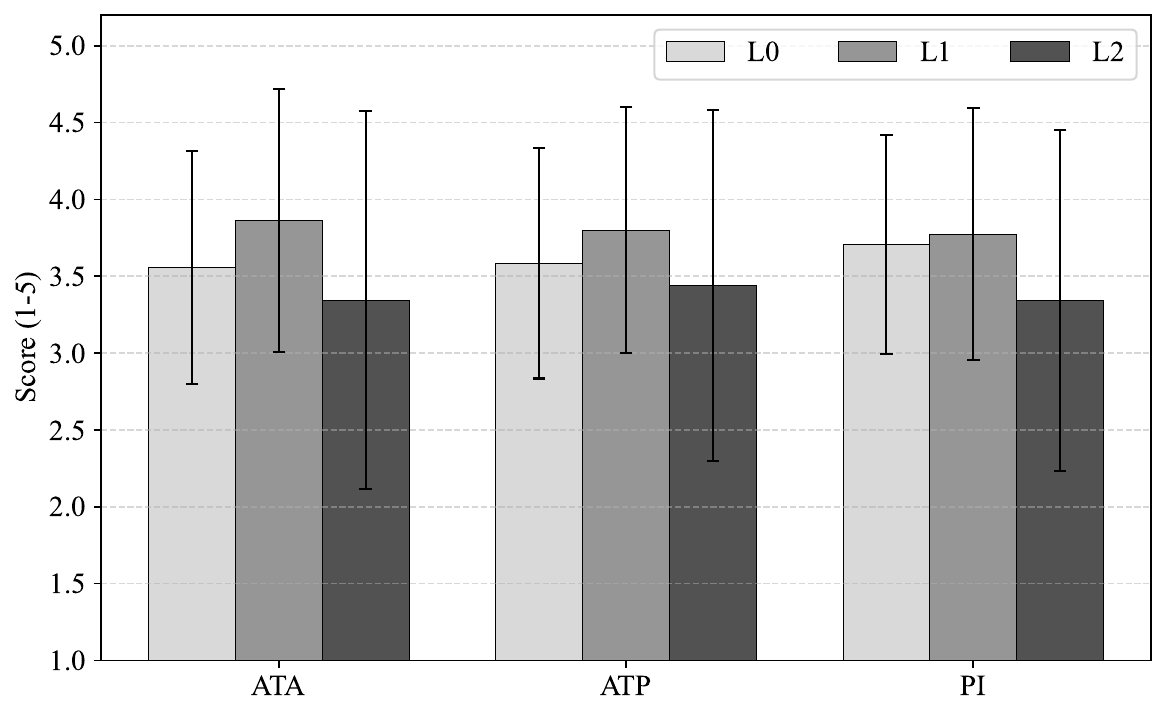}
  \end{center}
  \caption{Mean and standard deviation by construct and level of personalization.}
  \label{fig:barplot}
\end{figure}

We test our hypotheses using a repeated-measures ANOVA for H1 and planned one-tailed paired t-tests for H2 and H3. The effects of perceived personalization and perceived creepiness are also evaluated using a repeated-measures ANOVA. We test the sphericity assumption using Mauchly's test for all repeated-measures ANOVAs~\citep{mauchly1940significance}. When violated, we apply Greenhouse-Geisser corrections to the degrees of freedom~\citep{greenhousegeiser1959}. An overview of the means and standard deviations across levels and constructs is shown in \Cref{fig:barplot} with the exact values reported in \Cref{tab:descriptives}. For all three constructs results increase from L0 to L1 and decrease again at the highest level L2. Notably, the mean values at L2 are lower than the baseline L0 for each construct.

\paragraph{Hypothesis 1.}
The repeated-measures ANOVA reveal a significant effect of level of personalization on ATA, ATP, and PI (see \Cref{tab:h1-anova}), supporting H1 that levels of personalization differ in their effects for all constructs.

\begin{table}[ht]
\centering
\begin{tabular}{lrrr}
\toprule
Construct & $F(df_1, df_2)$ & p-value & $\eta^2_{GG}$ \\
\midrule
ATA & $\mathbf{16.68(1.46, 144.94)}$ & $\mathbf{< .001}$ & $\mathbf{.046}$ \\
ATP & $\mathbf{10.01(1.31, 130.11)}$ & $\mathbf{< .001}$ & $\mathbf{.026}$ \\
PI  & $\mathbf{17.84(1.31, 129.37)}$ & $\mathbf{< .001}$ & $\mathbf{.043}$ \\
\bottomrule
\end{tabular}
\caption{Repeated-measures ANOVA results for H1. Greenhouse-Geisser corrected degrees of freedom are reported due to sphericity violations in all constructs.}
\label{tab:h1-anova}
\end{table}

\paragraph{Hypothesis 2.}
We use two planned one-tailed paired t-tests to test whether any of the levels of personalization L1 or L2 exceed the baseline L0, for all three constructs. Our prediction is partially confirmed. The results support H2 for ATA and ATP. For both, L1 is significantly higher than L0 (see \Cref{tab:h2-ttests}). For PI we do not observe a significant increase of L1 over L0. For all three constructs L2 does not significantly exceed the baseline. In all three cases, the observed means for L2 were numerically lower than the baseline condition (see \Cref{tab:descriptives}), resulting in the non-significant one-tailed comparisons.

\paragraph{Hypothesis 3.}
In H3, we predict that the constructs, based on levels of personalization, follow an inverted-U pattern, with a moderate level of personalization (L1) outperforming extreme levels (L0, L2). To test for the inverted-U pattern, we use two planned one-tailed paired t-tests ($L1 > L0, L1 > L2$). Results provide mixed support: L1 significantly exceeds L0 for ATA and ATP, but not for PI---as already indicated in H2. For all three constructs, L1 is significantly greater than L2, as shown in \Cref{tab:h2-ttests}. Thus, H3 is supported only for ATA and ATP, as the inverted-U pattern is not observed for PI.

\begin{table}[ht]
\centering
\begin{tabular}{llrrrr}
\toprule
Construct & Comparison & $t(99)$ & $p$ & $d$ \\
\midrule
ATA & $L1 > L0$ & $\mathbf{5.28}$ & $\mathbf{< .001}$ & $\mathbf{.38}$ \\
    & $L2 > L0$ & $-1.98$         & $.975$            & $.21$          \\
    & $L1 > L2$ & $\mathbf{5.31}$ & $\mathbf{< .001}$ & $\mathbf{.49}$ \\
\midrule
ATP & $L1 > L0$ & $\mathbf{4.67}$ & $\mathbf{< .001}$ & $\mathbf{.28}$ \\
    & $L2 > L0$ & $-1.45$         & $.925$            & $.15$          \\
    & $L1 > L2$ & $\mathbf{4.18}$ & $\mathbf{< .001}$ & $\mathbf{.37}$ \\
\midrule
PI  & $L1 > L0$ & $1.38$ & $.085$     & $.09$ \\
    & $L2 > L0$ & $-3.69$         & $.999$              & $.39$          \\
    & $L1 > L2$ & $\mathbf{5.52}$ & $\mathbf{< .001}$  & $\mathbf{.44}$  \\
\bottomrule
\end{tabular}
\caption{Planned one-tailed paired t-tests for H2 ($level > L0$) and H3 ($L1 > L0, L1 > L2$). One-tailed p-values. H2 is supported if at least one comparison is significant per construct. H3 is supported if L1 is significantly greater than both L0 and L2.}
\label{tab:h2-ttests}
\end{table}

\paragraph{Exploratory analysis.}
A repeated measures ANOVA confirms that levels of personalization significantly affect perceived personalization, as well as perceived creepiness. For both, we test differences with post-hoc, Holm-Bonferroni corrected pairwise comparisons. As shown in \Cref{tab:iv_descriptive}, the tests reveal that each personalized condition is perceived as significantly more personalized than the previous (L2 > L1 > L0). The same behavior is observed for perceived creepiness. Each personalized condition is perceived as significantly more creepy than the previous. These results suggest that higher levels of personalization lead to greater perceived differences, but at the same time elicit greater concerns among participants.

\begin{table}[htb]
\centering
\begin{tabular}{llrrrr}
\toprule
Construct & Comparison & $t(99)$ & $p$ & $d$ \\
\midrule
Perceived personalization & $L1 > L0$ & $\mathbf{7.05}$ & $\mathbf{< .001}$ & $\mathbf{0.41}$ \\
                          & $L2 > L0$ & $\mathbf{5.97}$ & $\mathbf{< .001}$ & $\mathbf{0.67}$ \\
                          & $L2 > L1$ & $\mathbf{2.94}$ & $\mathbf{.004}$   & $\mathbf{0.28}$ \\
\midrule
Perceived creepiness      & $L1 > L0$ & $\mathbf{2.92}$ & $\mathbf{.004}$   & $\mathbf{0.23}$ \\
                          & $L2 > L0$ & $\mathbf{7.27}$ & $\mathbf{< .001}$ & $\mathbf{0.89}$ \\
                          & $L2 > L1$ & $\mathbf{6.96}$ & $\mathbf{< .001}$ & $\mathbf{0.71}$ \\
\bottomrule
\end{tabular}
\caption{Post-hoc pairwise comparisons (Holm-Bonferroni corrected p-values) for perceived personalization and perceived creepiness.}
\label{tab:iv_descriptive}
\end{table}

To better understand the interaction between levels of personalization and the different constructs, we conduct a parallel multilevel mediation analysis, using linear mixed-models, to examine the mechanisms through which levels of personalization affect the main constructs via the mediators perceived personalization and perceived creepiness.
Using cluster bootstrap ($5{,}000$ iterations) with random intercepts for participants, we estimate unique indirect effects for each level of personalization relative to baseline (L0). Path coefficients show that the levels of personalization (L1, L2) increase both perceived personalization and perceived creepiness, though creepiness does not show a significant effect on L1. We provide a concise overview in \Cref{tab:mediation_paths}. The two mediators, in turn, predict constructs in opposite directions, as shown in \Cref{tab:mediation_paths}. 

\begin{table}[ht]
\centering
\begin{tabular}{lrrr}
\toprule
Path & $b$ & SE & $p$ \\
\midrule
\multicolumn{4}{l}{\textit{Level of personalization $\rightarrow$ mediators}} \\
L1 $\rightarrow$ perceived personalization & $\mathbf{.47}$ & $\mathbf{.11}$ & $\mathbf{< .001}$ \\
L2 $\rightarrow$ perceived personalization & $\mathbf{.81}$ & $\mathbf{.11}$ & $\mathbf{< .001}$ \\
L1 $\rightarrow$ perceived creepiness & $.14$ & $.10$ & $.189$ \\
L2 $\rightarrow$ perceived creepiness & $\mathbf{.83}$ & $\mathbf{.10}$ & $\mathbf{< .001}$ \\
\midrule
\multicolumn{4}{l}{\textit{Mediators $\rightarrow$ outcomes}} \\
Perceived personalization $\rightarrow$ ATA & $\mathbf{.39}$ & $\mathbf{.04}$ & $\mathbf{< .001}$ \\
Perceived creepiness $\rightarrow$ ATA & $\mathbf{-.56}$ & $\mathbf{.04}$ & $\mathbf{< .001}$ \\
Perceived personalization $\rightarrow$ ATP & $\mathbf{.35}$ & $\mathbf{.03}$ & $\mathbf{< .001}$ \\
Perceived creepiness $\rightarrow$ ATP & $\mathbf{-.49}$ & $\mathbf{.04}$ & $\mathbf{< .001}$ \\
Perceived personalization $\rightarrow$ PI & $\mathbf{.31}$ & $\mathbf{.03}$ & $\mathbf{< .001}$ \\
Perceived creepiness $\rightarrow$ PI & $\mathbf{-.49}$ & $\mathbf{.04}$ & $\mathbf{< .001}$ \\
\bottomrule
\end{tabular}
\caption{Path coefficients of the parallel multilevel mediation model. The upper panel reports the effects of level of personalization on the mediators, and the lower panel reports the effects of the mediators on the outcome constructs.}
\label{tab:mediation_paths}
\end{table}

Indirect effects reveal a critical nonlinear pattern: At L1, perceived personalization and perceived creepiness produce a net positive indirect effect for ATA and ATP, as connection benefits outweigh creepiness costs, while the indirect effect for PI at L1 and the indirect effects across all constructs at L2 are not significantly different from zero. These results are driven by strongly increasing creepiness perceptions that overwhelm personalization gains (see \Cref{tab:mediation_effects}). The total effects confirm a peak for the moderate level of personalization. ATA and ATP are increased at L1 relative to L0, while PI does not reach statistical significance. At L2, ATA and PI decrease, while the effect is not significant for ATP. A concise overview of these results is presented in \Cref{tab:mediation_effects}. These findings demonstrate the personalization-paradox in our setting and show that moderate personalization enhances outcomes through a feeling of personal relevance but higher levels trigger creepiness reactions that produce net negative effects.
\section{Discussion}
\label{sec:discussion}

\begin{figure}[htbp]
    \centering
    \begin{minipage}[t]{0.24\textwidth}
        \centering
        \includegraphics[width=\textwidth]{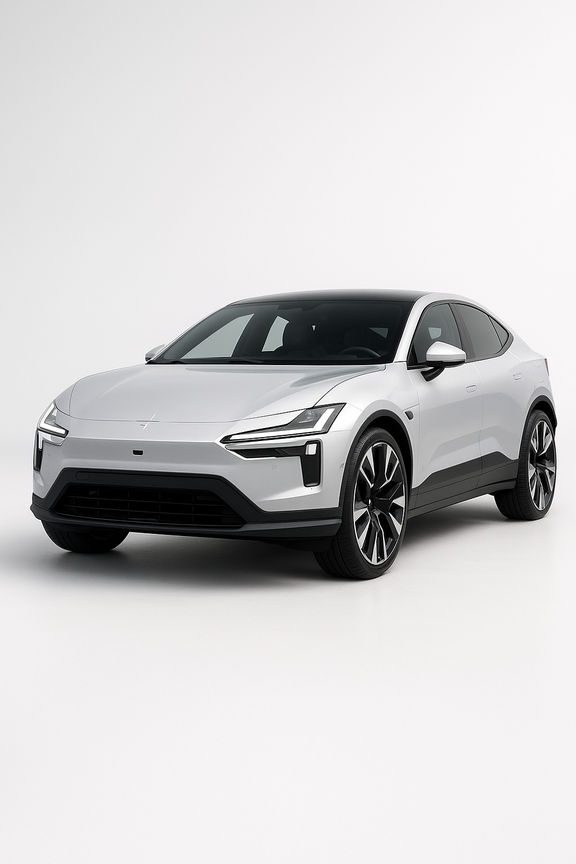}
    \end{minipage}
    \hfill
    \begin{minipage}[t]{0.24\textwidth}
        \centering
        \includegraphics[width=\textwidth]{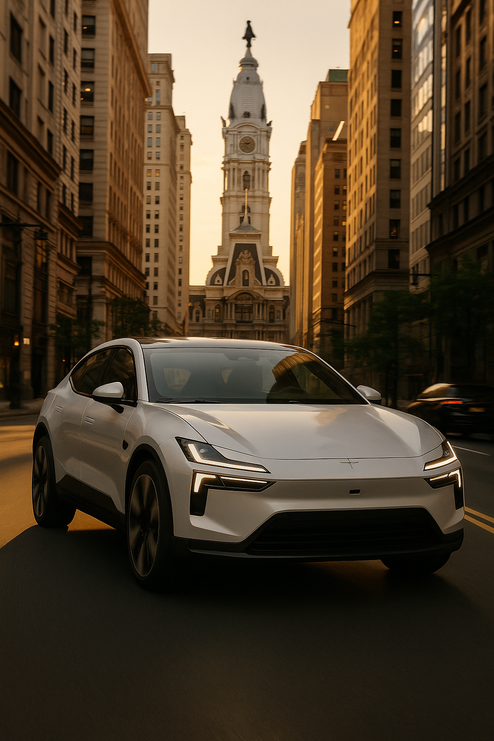}
    \end{minipage}
    \hfill
    \begin{minipage}[t]{0.24\textwidth}
        \centering
        \includegraphics[width=\textwidth]{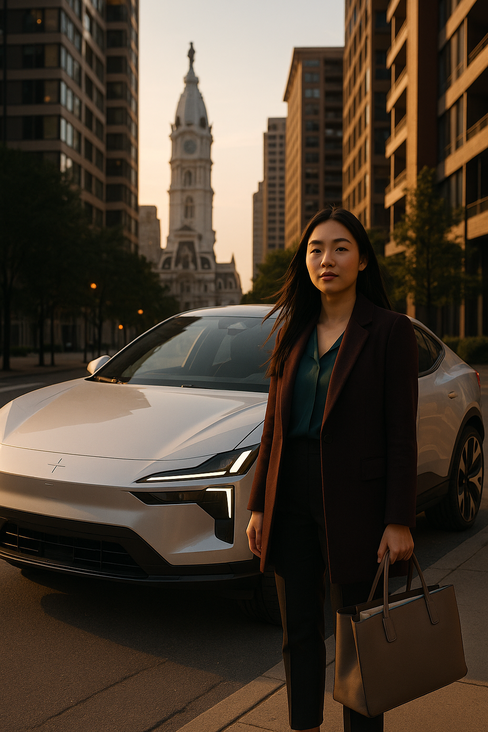}
    \end{minipage}
    
    \medskip
    
    \begin{minipage}[t]{0.24\textwidth}
        \centering
        L0
    \end{minipage}
    \hfill
    \begin{minipage}[t]{0.24\textwidth}
        \centering
        L1
    \end{minipage}
    % \hfill
    % \begin{minipage}[t]{0.24\textwidth}
    %     \centering
    %     L2
    % \end{minipage}
    \hfill
    \begin{minipage}[t]{0.24\textwidth}
        \centering
        L2
    \end{minipage}
    
    \caption{From left to right: Showcase of different levels of personalization, L0--L2, for the same participant persona and product.}
    \label{fig:example_images_level}
\end{figure}

This study examined whether GenAI can generate personalized advertising imagery from customer data and how different levels of personalization affect ATA, ATP, and PI. The results demonstrate the feasibility of the framework and show that moderate personalization (L1) improves ATA and ATP, whereas stronger personalization (L2) does not add benefit and can reduce outcomes. For ATA and ATP, the results therefore indicate an inverted-U pattern. The framework also successfully produces visibly different levels of  personalization, as illustrated in \Cref{fig:example_images_level} and \Cref{fig:image_grid_all}.

The mediation analysis shows that perceived personalization and perceived creepiness operate as competing mechanisms. Perceived personalization positively predicts ATA, ATP, and PI, whereas perceived creepiness has a slightly stronger negative effect. At L1, personalization benefits outweigh creepiness costs for ATA and ATP. At L2, however, increasing creepiness neutralizes these benefits across all three constructs. This pattern supports the personalization-paradox in the context of AI-generated advertising imagery and suggests that stronger visual personalization can undermine its own relevance benefits.

Notably, the decline from L1 to L2 cannot be explained by perceived creepiness alone. As shown in \Cref{tab:mediation_effects}, the direct effects of personalization also diminish sharply: the positive direct effects of L1 disappear at L2 for ATA and ATP, while PI even shows a negative direct effect at L2. One explanation could be reduced image quality at higher personalization. Additional customer information may increase prompt complexity and lead to ``catastrophic neglect'', where the model fails to represent parts of the image correctly~\citep{chefer2023attend,fu2024enhancing}. Distorted features, implausible scenes, or other visual artifacts may therefore reduce advertising effectiveness and simultaneously increase perceived creepiness. Beyond these technical limitations, stronger personalization may also increase algorithm aversion, as customers can be less willing to rely on algorithms for subjective tasks~\citep{castelo2019task}. Thus, the negative effect of strong personalization likely reflects the personalization-paradox, current technical limitations of GenAI-based image generation, and algorithm aversion in a subjective advertising context. Future implementations may require more selective customer data use, specialized models, or automated quality-control mechanisms before deployment.

The fact that H1 and H2 are supported only for ATA and ATP, but not for PI, is consistent with a hierarchy-of-effects perspective. While personalization affects earlier attitudinal responses, such as ATA and ATP, the effect does not extend to PI, which represents a more downstream and behavior-proximal outcome \citep{lavidge1961model}.

\paragraph{Theoretical implications.}
Our findings contribute to several areas of research. First, we extend the research on personalized marketing, specifically the topics of personalized advertising and DCO.
Our results also clarify the contribution of this study relative to the growing literature on AI-generated advertising. Prior work has shown that synthetic advertising represents a new generation of advertising manipulation enabled by AI~\citep{campbell2022preparing}, and that customer responses to AI-generated advertisements can be negative when the use of AI conflicts with brand expectations~\citep{to2025ai}. We extend this literature by shifting attention away from comparison of AI-generated and human-created advertisements to the degree of personalization within AI-generated visual content. This distinction is theoretically important because personalization depth introduces a second evaluative tension: the same customer data that can increase relevance may also increase creepiness and thereby reduce ATA, ATP, and PI.
Previous research in the field of DCO focuses mainly on generating personalized advertising text with AI~\citep{matz2024potential,patil2024generative}. Different marketing slogans or image descriptions are generated for individual customers based on their data. Other work replaces parts of the images with GenAI, based on context~\citep{yang2024new,czapp2024dynamic}. Our study goes beyond this approach. We fully generate the imagery with AI, given a product, and include high levels of personalization, based on customer data. We embed customer preferences and portrayals into the product to improve personalization. We show that this approach can yield positive results on ATA and ATP, thereby confirming the general feasibility with our technical framework. In line with the broader argument that GenAI enables novel personalization interventions by making large-scale creative personalization technically feasible~\citep{lemmens2025personalization}, our results demonstrate that personalization can be enacted directly at the level of fully generated visuals rather than being limited to modular asset recombination or partial edits. Previous research in personalized advertising finds that personalized advertising improves customer attention and interaction, as well as increasing purchases~\citep{matz2017psychological,matz2024potential}, and we confirm a similar behavior with fully AI-generated content.
In addition, our findings contribute to the literature on the privacy-paradox~\citep{aguirre2015unraveling, chandra2022personalization,cloarec2020personalization,aguirre2016personalization,goldfarb2011online} in the domain of fully AI-generated advertising imagery. The observed behavior is a trade-off between shared information and gains in personalization. Customers weigh intrusiveness in their personal data against the added value from personalized advertisements. We observe the same behavior for our AI-generated images. However, we note that the diminishing returns we observe in our study could also stem from different factors, such as image quality.

\begin{figure}[htbp]
    \centering
    \begin{minipage}[t]{0.24\textwidth}
        \centering
        \includegraphics[width=\textwidth]{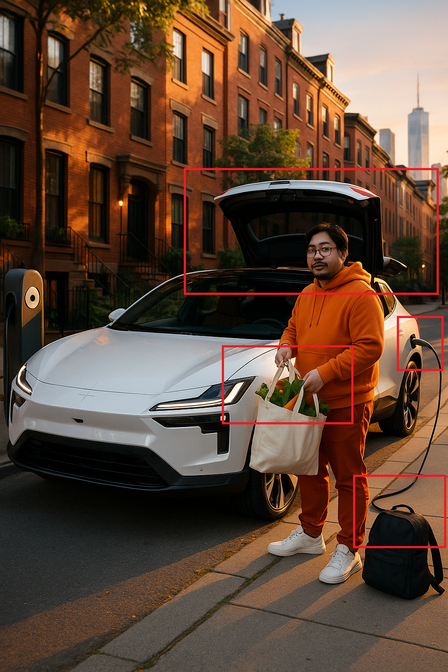}
    \end{minipage}
    \hfill
    \begin{minipage}[t]{0.24\textwidth}
        \centering
        \includegraphics[width=\textwidth]{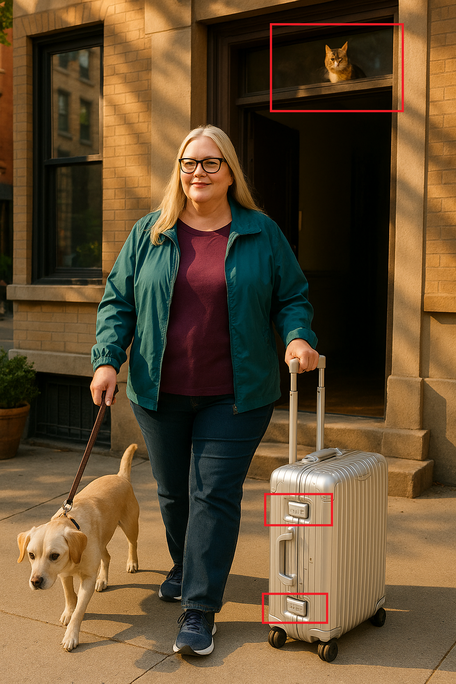}
    \end{minipage}
    \hfill
    \begin{minipage}[t]{0.24\textwidth}
        \centering
        \includegraphics[width=\textwidth]{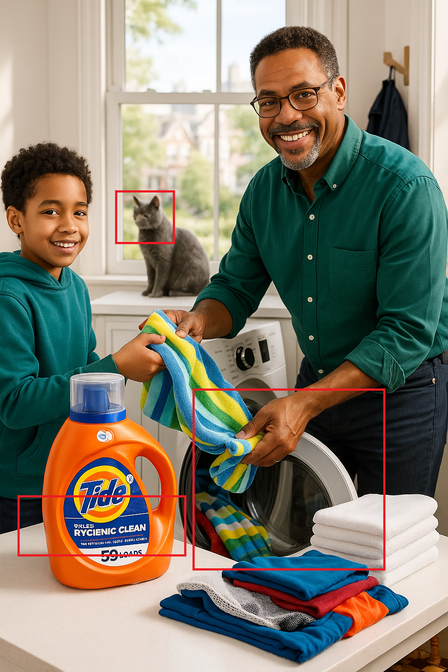}
    \end{minipage}
    \hfill
    \begin{minipage}[t]{0.24\textwidth}
        \centering
        \includegraphics[width=\textwidth]{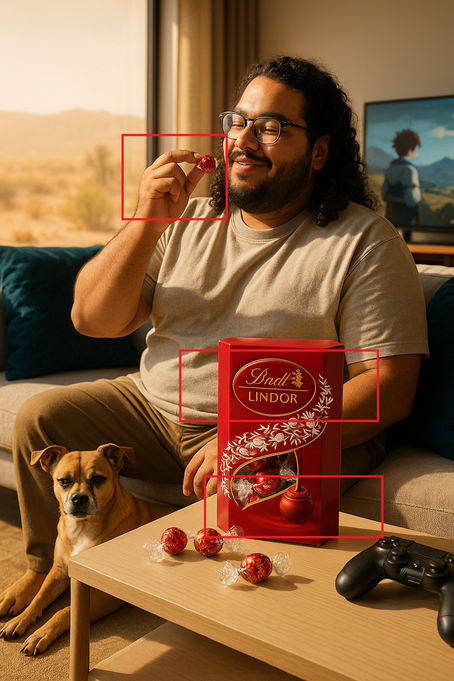}
    \end{minipage}
    
    \medskip
    
    \begin{minipage}[t]{0.24\textwidth}
        \centering
        Electric vehicle
    \end{minipage}
    \hfill
    \begin{minipage}[t]{0.24\textwidth}
        \centering
        Luxury suitcase
    \end{minipage}
    \hfill
    \begin{minipage}[t]{0.24\textwidth}
        \centering
        Laundry detergent
    \end{minipage}
    \hfill
    \begin{minipage}[t]{0.24\textwidth}
        \centering
        Chocolates
    \end{minipage}
    
    \caption{Some images contain obvious flaws such as the reversed trunk on the electric vehicle, misplaced or distorted regions for the luxury suitcase or text-generation errors in the laundry detergent and chocolate logo. All displayed images are generated with the strong level of personalization (L2) over different products, with major flaws highlighted in red.}
    \label{fig:example_images_bad}
\end{figure}

\paragraph{Practical implications.}
Targeting customers with data-driven advertisements is already standard practice~\citep{chandra2022personalization,murray2009personalization,naumov2019deep}, fueled by the growth of AI and the abundance of data available from customer interactions or directly from social media. This development is part of a broader shift in marketing practice driven by new digital technologies that change how firms create, deliver, and optimize marketing value~\citep{hoffman2022rise}. While marketers have become excellent at matching the right product to the right customer, personalizing the actual creative content of an advertisement remains expensive. Usually, brands will limit themselves to a small handful of advertisement variations because the production cost of advertisements, especially advertising imagery, is high~\citep{hartmann2025power}. GenAI eliminates this barrier, allowing for the automated creation of advertisement assets at scale. Research shows that AI-generated advertisements can now match or even outperform human-made ones~\citep{hartmann2025power}.
This presents a marketing innovation~\citep{hoffman2022rise} with new opportunities for personalized marketing: Companies can now tailor both content and presentation to individual interests and personalities without the high overhead.
Our study shows that moderate personalization is well-received, allowing marketers to optimize both the recommendation and the creative content. Instead of only focusing on finding the right audience, marketers can now simultaneously tailor the advertisement itself to that audience, creating a far more precise alignment between the product and the customer.
This development also has implications for platforms. We argue that a bottleneck in current personalization efforts is modest quality of AI-generated images, which diminishes the potential returns. We present a selection of faulty images in \Cref{fig:example_images_bad}.
However, if these technical challenges are solved, the gains could be significantly higher. Platforms such as Instagram or Facebook sit at the intersection of marketers and customers. They own the data, host the advertisements, and have an active user base. This positions platform providers to offer AI-generated personalized advertisements as a core service. They have the infrastructure and the technical expertise to build these systems, allowing them to move beyond simply being the channel for advertisement distribution. By leveraging their existing data infrastructure and building their own advertising models, they can offer a new service that manages both parts of the match---automatically building the right advertisement for the right customer the moment it is shown. This allows platforms to move from simple distribution to actively optimizing every variable to find the best possible result for every customer. In summary, in the future, these platforms can leverage customer data for two decisions: which product to present to which customer, and what level of personalization they choose to have the optimal result.

\paragraph{Future research.}
Future research should refine the understanding of personalization in AI-generated advertising imagery in four directions. First, studies should isolate individual customer attributes, such as demographic, lifestyle, or appearance-related cues, to identify which inputs increase perceived personalization without increasing perceived creepiness or reducing ATA, ATP, and PI. Second, future work should examine how directly customers should be visually represented in personalized advertisements, ranging from abstract lifestyle cues to visually similar persons or direct customer depictions. This would clarify how customer likeness affects perceived personalization, perceived creepiness, ATA, ATP, and PI.
Third, research should investigate boundary conditions under which deeper personalization becomes beneficial or harmful. Our findings suggest that stronger personalization can reduce marketing outcomes, consistent with the personalization-paradox~\citep{aguirre2015unraveling,chandra2022personalization}. Yet this effect may depend on data-use expectations, perceived appropriateness, platform context, disclosure, consent mechanisms, and product category~\citep{aguirre2015unraveling,de2022personalization}. Finally, future work should separate the effects of personalization depth from image quality. Higher personalization may increase generation complexity and produce artifacts, which can reduce ATA, ATP, and PI and increase perceived creepiness. Systematic quality assessments, controlled artifact manipulations, and automated quality-control mechanisms could improve the reliability of personalized image generation.

% Ethics / Fair use Research?
\paragraph{Ethical considerations.}
AI-generated personalized advertising raises ethical concerns related to privacy, trust, and customer autonomy \citep{hermann2022leveraging,hermann2022artificial}. First, personalization depends on customer data, yet customers may not be fully aware of the extent to which such data are collected, processed, or inferred. This challenges informed consent, especially when customers lack the technical understanding needed to evaluate how their data are used \citep{zettler2025personalized}. AI-based marketing can further intensify privacy risks when data are collected without sufficient consent, leaked or de-anonymized, or used to infer sensitive attributes from customer and interaction data \citep{hermann2022leveraging}.
Second, AI-generated advertising may erode trust when generated content misrepresents products or brand promises. Hallucinated, exaggerated, or overly polished product presentations can create expectations that do not match reality. Similarly, visual flaws in generated advertising imagery may negatively affect brand perception. Both cases can damage the brand--customer relationship by undermining credibility.
Third, GenAI-based personalization can threaten customer autonomy by shaping choice architectures in highly persuasive or manipulative ways~\citep{hermann2022artificial}. Personalized advertising may bypass reflective decision-making when emotionally engaging content is optimized toward persuasion rather than informed choice, potentially contributing to unnecessary or uncontrolled spending~\citep{matz2024potential,zettler2025personalized}. To address these risks, marketers and platforms should implement meaningful transparency and governance mechanisms that support intelligibility, accountability, and appropriate human oversight in ethically salient marketing contexts~\citep{hermann2022artificial}.
\section{Conclusion}
\label{sec:conclusion}

This paper examines whether GenAI enables scalable creative personalization in advertising imagery and whether increasing personalization improves marketing outcomes. We address this gap by proposing and implementing a fully automated framework that transforms real customer data into personalized advertising imagery through a three-step pipeline (persona creation, prompt generation with an LLM, and image synthesis with a T2I model). We operationalize personalization across three levels (L0--L2) that systematically vary the amount and specificity of user information, enabling a controlled test of personalization depth in AI-generated visuals.
In a preregistered two-stage within-subjects study with $N=100$ participants and four products, we find that participants reliably perceive differences in personalization across conditions, confirming that GenAI incorporates customer data into generated imagery. Personalization affects ATA, ATP, and PI, but the effect is not monotonic. Light personalization (L1) yields the highest descriptive values for ATA, ATP, and PI and significantly outperforms the non-personalized baseline (L0) for ATA and ATP. In contrast, strong personalization (L2) does not improve outcomes and can reduce them.
To explain these diminishing returns, we show that perceived personalization and perceived creepiness operate as competing mechanisms. Higher levels of personalization increase perceived personalization, which positively predicts ATA, ATP, and PI, but they also increase perceived creepiness, which negatively predicts the three measures and grows more strongly at high personalization. The resulting net effect favors L1, where connection benefits outweigh creepiness costs, whereas high personalization shifts the balance toward less favorable reactions. These findings extend the personalization-paradox to fully AI-generated advertising imagery and provide actionable guidance: effective GenAI-driven image personalization is achievable, but it benefits from conservative use of personal cues.
Future research should refine personalization strategies at the attribute level, improve robustness and visual fidelity under richer profile inputs, and investigate boundary conditions such as product types, disclosure strategies, and platform contexts to determine when deeper personalization becomes acceptable and beneficial.
\section*{Declaration of generative AI and AI-assisted technologies in the manuscript preparation process}

During the preparation of this work, the authors used ChatGPT, Claude, Gemini and DeepL Write in order to assist with language refinement and clarity. After using these tools, the authors reviewed and edited the content as needed and take full responsibility for the content of the published article.

\clearpage

\appendix
\section{Appendix}
\label{sec:appendix}

\begin{table}[ht]
\centering
\begin{tabular}{lrrrrrr}
\toprule
 & L0 & & L1 & & L2 & \\
\cline{2-3} \cline{4-5} \cline{6-7}
 & $M$ & $SD$ & $M$ & $SD$ & $M$ & $SD$ \\
\midrule
ATA & 3.56 & .76 & 3.86 & .85 & 3.35 & 1.23 \\
ATP & 3.59 & .75 & 3.80 & .80 & 3.44 & 1.14 \\
PI  & 3.71 & .71 & 3.78 & .82 & 3.34 & 1.11 \\
\bottomrule
\end{tabular}
\caption{Descriptive statistics by construct and level of personalization.}
\label{tab:descriptives}
\end{table}

\begin{table}[ht]
\centering
\small
\begin{tabular}{lllrr}
\toprule
 & Contrast & Effect & Coef. & 95\% CI \\
\midrule
ATA & $L1$ vs $L0$ & Indirect via perceived personalization & $\mathbf{.18}$ & $[.13, .25]$ \\
    &              & Indirect via perceived creepiness & $\mathbf{-.08}$ & $[-.14, -.03]$ \\
    &              & Total indirect  & $\mathbf{.10}$ & $[.02, .19]$ \\
    &              & Direct ($c'$)   & $\mathbf{.20}$ & $[.10, .31]$ \\
    &              & Total ($c$)     & $\mathbf{.31}$ & $[.20, .42]$ \\
\addlinespace
    & $L2$ vs $L0$ & Indirect via perceived personalization & $\mathbf{.31}$ & $[.21, .43]$ \\
    &              & Indirect via perceived creepiness & $\mathbf{-.46}$ & $[-.63, -.30]$ \\
    &              & Total indirect  & $-.15$ & $[-.35, .05]$ \\
    &              & Direct ($c'$)   & $-.07$ & $[-.26, .12]$ \\
    &              & Total ($c$)     & $\mathbf{-.21}$ & $[-.43, -.002]$ \\
\midrule
ATP & $L1$ vs $L0$ & Indirect via perceived personalization & $\mathbf{.17}$ & $[.11, .23]$ \\
    &              & Indirect via perceived creepiness & $\mathbf{-.07}$ & $[-.13, -.02]$ \\
    &              & Total indirect  & $\mathbf{.10}$ & $[.02, .18]$ \\
    &              & Direct ($c'$)   & $\mathbf{.12}$ & $[.03, .21]$ \\
    &              & Total ($c$)     & $\mathbf{.22}$ & $[.13, .31]$ \\
\addlinespace
    & $L2$ vs $L0$ & Indirect via perceived personalization & $\mathbf{.28}$ & $[.19, .39]$ \\
    &              & Indirect via perceived creepiness & $\mathbf{-.40}$ & $[-.57, -.26]$ \\
    &              & Total indirect  & $-.12$ & $[-.31, .07]$ \\
    &              & Direct ($c'$)   & $-.03$ & $[-.20, .14]$ \\
    &              & Total ($c$)     & $-.15$ & $[-.35, .05]$ \\
\midrule
PI  & $L1$ vs $L0$ & Indirect via perceived personalization & $\mathbf{.14}$ & $[.09, .20]$ \\
    &              & Indirect via perceived creepiness & $\mathbf{-.07}$ & $[-.13, -.02]$ \\
    &              & Total indirect  & $.07$ & $[-.01, .15]$ \\
    &              & Direct ($c'$)   & $-.01$ & $[-.11, .08]$ \\
    &              & Total ($c$)     & $.07$ & $[-.03, .16]$ \\
\addlinespace
    & $L2$ vs $L0$ & Indirect via perceived personalization & $\mathbf{.25}$ & $[.16, .35]$ \\
    &              & Indirect via perceived creepiness & $\mathbf{-.40}$ & $[-.57, -.26]$ \\
    &              & Total indirect  & $-.16$ & $[-.35, .02]$ \\
    &              & Direct ($c'$)   & $\mathbf{-.21}$ & $[-.38, -.05]$ \\
    &              & Total ($c$)     & $\mathbf{-.37}$ & $[-.56, -.17]$ \\
\bottomrule
\end{tabular}
\caption{Mediation effects by level of personalization. Values are standardized effects with 95\% bootstrap confidence intervals. Bold values in the coefficient column indicate confidence intervals excluding zero.}
\label{tab:mediation_effects}
\end{table}

\begin{figure}[htb]
\centering
\setlength{\tabcolsep}{2pt}
\renewcommand{\arraystretch}{1.0}

\begin{adjustbox}{max width=\linewidth}
\begin{tabular}{@{}l c c c c@{}}
L0 &
\includegraphics[width=0.145\linewidth]{figures/image_prod-1_L0.png} &
\includegraphics[width=0.145\linewidth]{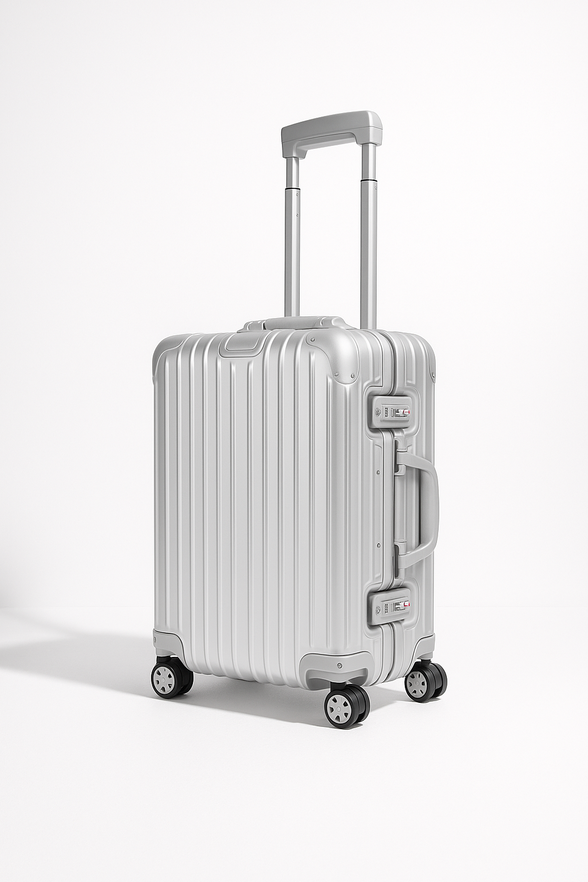} &
\includegraphics[width=0.145\linewidth]{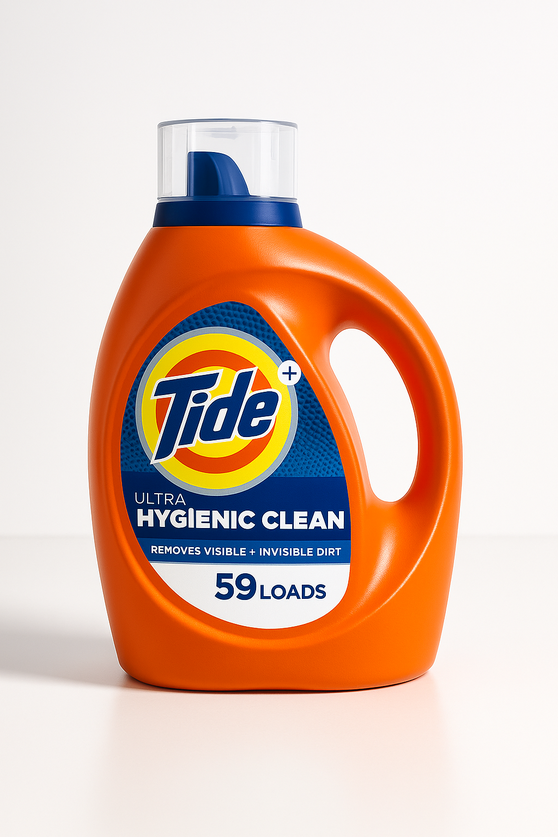} &
\includegraphics[width=0.145\linewidth]{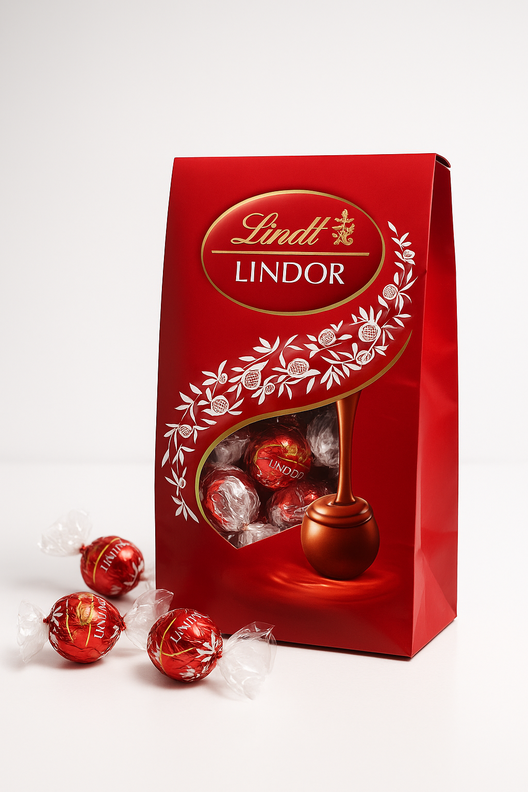} \\
L1 &
\includegraphics[width=0.145\linewidth]{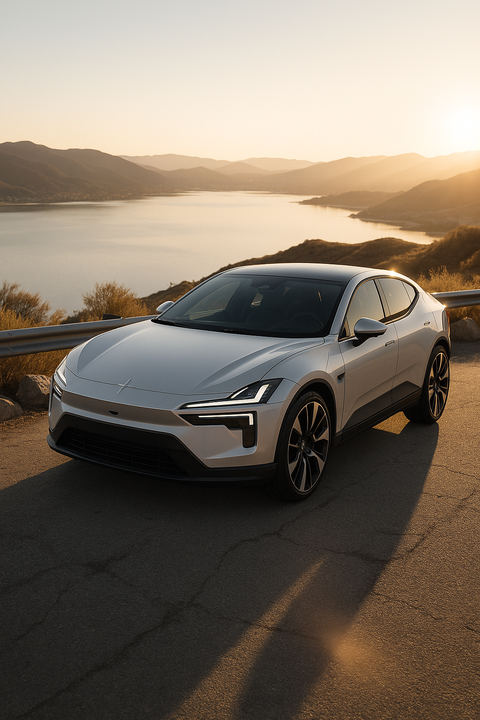} &
\includegraphics[width=0.145\linewidth]{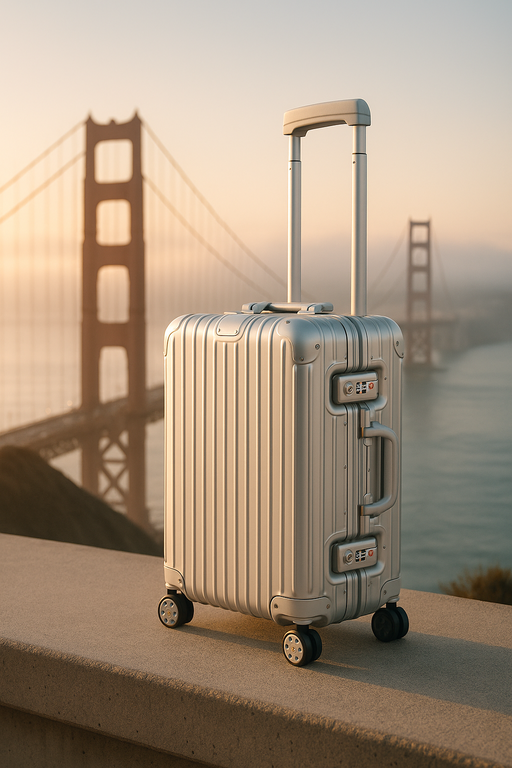} &
\includegraphics[width=0.145\linewidth]{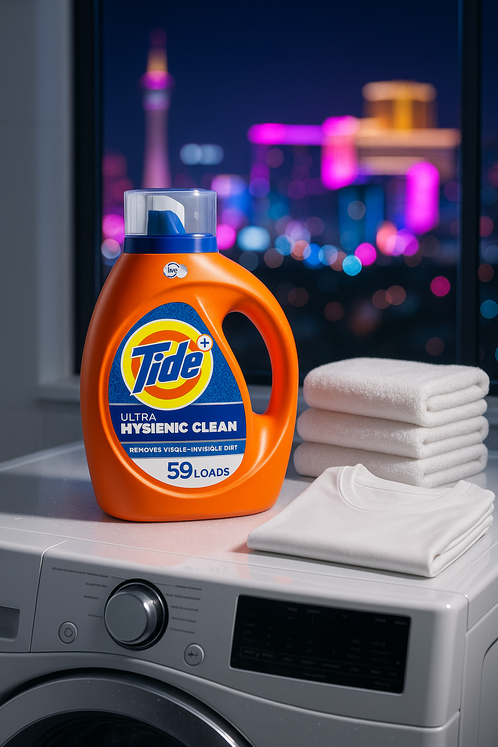} &
\includegraphics[width=0.145\linewidth]{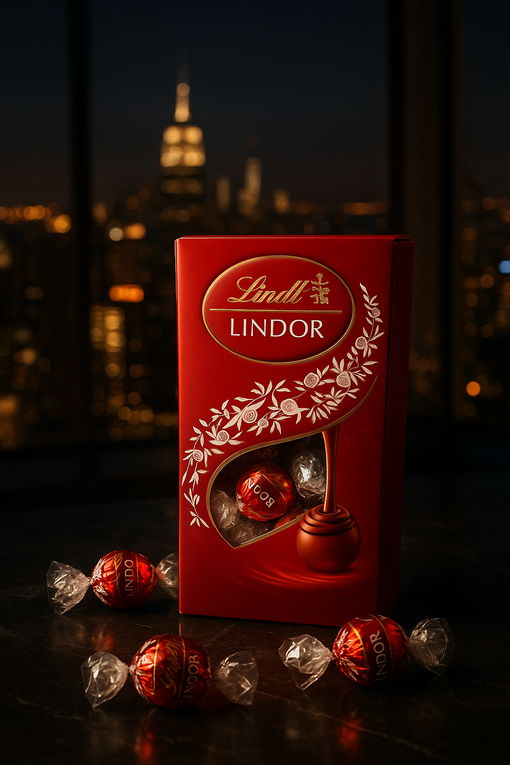} \\
L2 &
\includegraphics[width=0.145\linewidth]{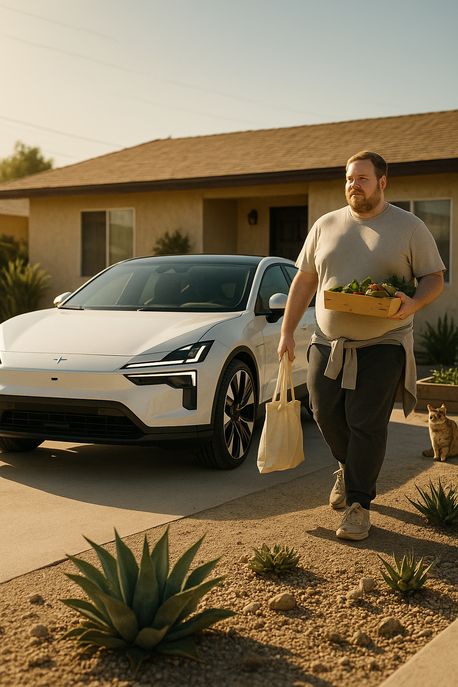} &
\includegraphics[width=0.145\linewidth]{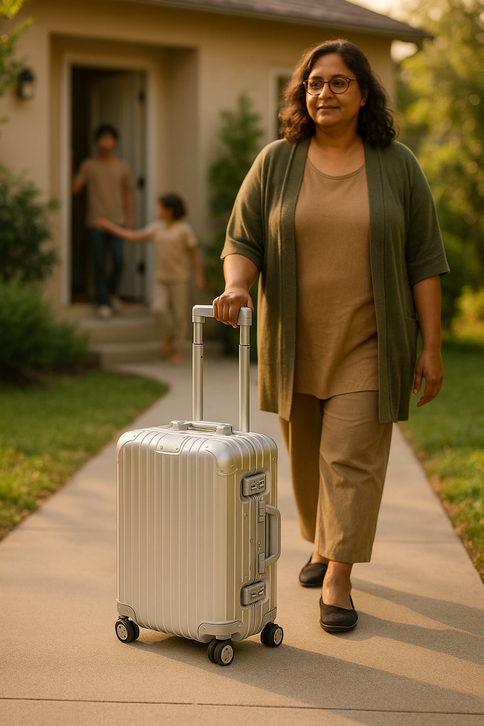} &
\includegraphics[width=0.145\linewidth]{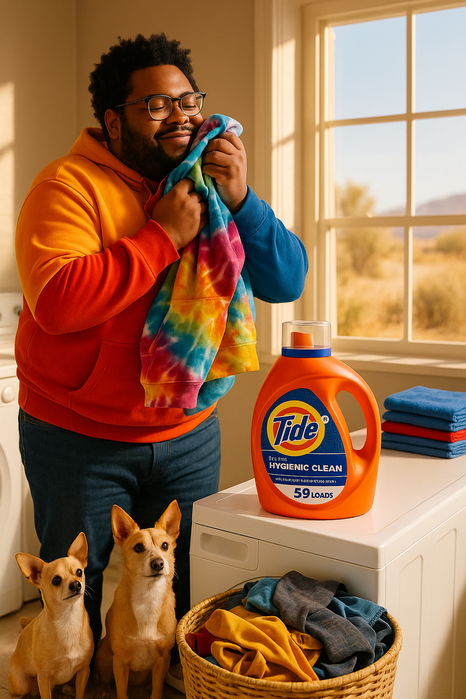} &
\includegraphics[width=0.145\linewidth]{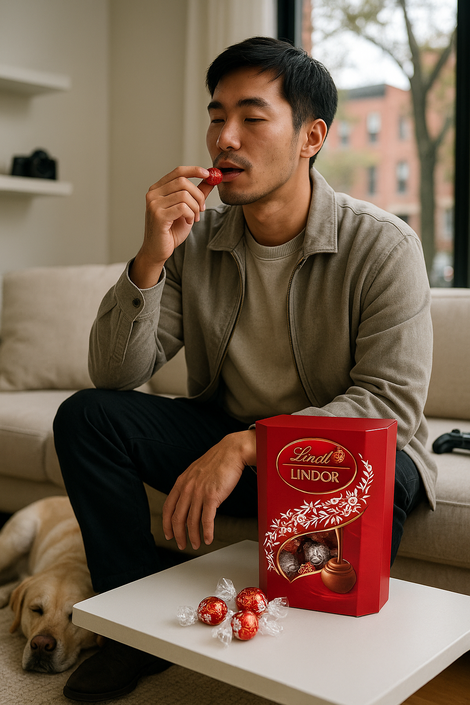} \\
\addlinespace[2pt]
& \shortstack{Electric\\vehicle}
& \shortstack{Luxury\\suitcase}
& \shortstack{Laundry\\detergent}
& \shortstack{Chocolates\\ \strut} \\
\end{tabular}
\end{adjustbox}

\caption{Example advertisements for all four products (columns) across levels of personalization (rows), generated from different participant personas.}
\label{fig:image_grid_all}
\end{figure}

\clearpage

\bibliographystyle{ACM-Reference-Format}
\bibliography{references}  %%% Uncomment this line and comment out the ``thebibliography'' section below to use the external .bib file (using bibtex) .




% 	\bibitem{keshet2016prediction}
% 	Keshet, Renato, Alina Maor, and George Kour.
% 	\newblock Prediction-Based, Prioritized Market-Share Insight Extraction.
% 	\newblock In {\em Advanced Data Mining and Applications (ADMA), 2016 12th International 
%                       Conference of}, pages 81--94,2016.

% \end{thebibliography}

\end{document}